\pdfoutput=1
\documentclass[11pt]{article}

\usepackage[final]{acl}

\usepackage{times}
\usepackage{amsmath}
\usepackage{latexsym}
\usepackage{cleveref}
\usepackage{array}

\usepackage[T1]{fontenc}
\usepackage[utf8]{inputenc}

\usepackage{microtype}

\usepackage{inconsolata}

\usepackage{graphicx}

\usepackage{amssymb}
\usepackage{multirow}
\usepackage{booktabs}
\usepackage{subcaption}
\usepackage{tcolorbox}
\usepackage{pifont}
\usepackage{xcolor}
\definecolor{c1}{cmyk}{0,0.6175,0.8848,0.1490} 
\definecolor{c2}{cmyk}{0.1127,0.6690,0,0.4431} 
\definecolor{c3}{rgb}{0.55, 0.71, 0.0}
\definecolor{c4}{cmyk}{0.6765,0.2017,0,0.0667} 
\definecolor{c5}{cmyk}{0,0.8765,0.7099,0.3647} 
\definecolor{forestgreen}{RGB}{34,139,34}
\definecolor{brickred}{RGB}{203,65,84}

\newtcbox{\hlprimarytab}{on line, rounded corners, box align=base, colback=red!10,colframe=white,size=fbox,arc=3pt, before upper=\strut, top=-2pt, bottom=-4pt, left=-2pt, right=-2pt, boxrule=0pt}
\newtcbox{\hlsecondarytab}{on line, box align=base, colback=c3!10,colframe=white,size=fbox,arc=3pt, before upper=\strut, top=-2pt, bottom=-4pt, left=-2pt, right=-2pt, boxrule=0pt}
\newtcbox{\hlmedium}{on line, box align=base, colback=gray!10,colframe=white,size=fbox,arc=3pt, before upper=\strut, top=-2pt, bottom=-4pt, left=-2pt, right=-2pt, boxrule=0pt}
\newcommand{\dashifted}{\raisebox{0.5\depth}{\tiny$\downarrow$}}

\newcommand{\da}[1]{{\scriptsize\hlprimarytab{\dashifted{#1}}}}

\newcommand{\syntheval}[0]{{\textsc{SynthTextEval}}}

\title{SynthTextEval: Synthetic Text Data Generation and Evaluation for High-Stakes Domains}

\author{
  Krithika Ramesh$^{\spadesuit}$ \quad Daniel Smolyak$^{\diamondsuit}$ \quad Zihao Zhao$^{\spadesuit}$ \\
  \textbf{Nupoor Gandhi}$^{\clubsuit}$ \quad \textbf{Ritu Agarwal}$^{\spadesuit}$  \quad \textbf{Margrét Bjarnadóttir}$^{\diamondsuit}$ \quad \textbf{Anjalie Field}$^{\spadesuit}$\\
  $^{\spadesuit}$Johns Hopkins University \\ $^{\diamondsuit}$University of Maryland, College Park \qquad $^{\clubsuit}$Carnegie Mellon University \\
        {\tt \{kramesh3, zzhao71, ritu.agarwal, anjalief\}@jhu.edu} \\ {\tt \{dsmolyak, mbjarnad\}@umd.edu} \\ {\tt{nmgandhi@cs.cmu.edu}}
}

\begin{document}
\maketitle
\begin{abstract}

We present \syntheval, a toolkit for conducting comprehensive evaluations of synthetic text. The fluency of large language model (LLM) outputs has made synthetic text potentially viable for numerous applications, such as reducing the risks of privacy violations in the development and deployment of AI systems in high-stakes domains.
Realizing this potential, however, requires principled consistent evaluations of synthetic data across multiple dimensions: its utility in downstream systems, the fairness of these systems, the risk of privacy leakage, general distributional differences from the source text, and qualitative feedback from domain experts. \syntheval{} allows users to conduct evaluations along all of these dimensions over synthetic data that they upload or generate using the toolkit's generation module. While our toolkit can be run over any data, we highlight its functionality and effectiveness over datasets from two high-stakes domains: healthcare and law.
By consolidating and standardizing evaluation metrics, we aim to improve the viability of synthetic text, and in-turn, privacy-preservation in AI development.
\end{abstract}

\section{Introduction}

Currently, in contexts involving sensitive data like healthcare and social services, accessing domain-specific text to develop AI systems and assistive tools necessitates strict data-sharing agreements and well-defined protocols to reduce privacy risks and maintain compliance with regulatory standards \cite{de2023guide, alberto2023impact}. Even with careful protocols, privacy-preservation is a key challenge, particularly if systems that have been developed using sensitive data are accessible to the public. While text anonymization can be used to redact or replace private identifiers from text, there is still significant risk of sensitive attributes in the data being de-identified or exposed \cite{staab24beyond, xin2024a, pang2024reconstructiondifferentiallyprivatetext}. 

Synthetic data offers a possible alternative means of privacy-preservation.
 If synthetic text is sufficiently reflective of real text, it could facilitate the development of tools or conducting analyses, while minimizing the risk of a privacy breach.
Beyond privacy-preservation, there has been a substantial amount of interest in other synthetic text applications, including fine-tuning task-specific models, serving as a proxy for human evaluations and auditing models  \cite{mitra2023orca2teachingsmall, xu2023wizardlmempoweringlargelanguage, tan-etal-2024-large, huang-etal-2023-large, min-etal-2023-factscore}. Although approaches for generating synthetic text may differ depending on the intended use-case, they typically share the same underlying objective - to produce diverse, high-quality data with patterns and signals that mimic the characteristics of real-world data distributions \cite{wu2024unigenunifiedframeworktextual, liu2024best}.

%There has been extensive research on generating synthetic data in a manner that enforces privacy guarantees by incorporating privacy-preserving elements. While some studies focus on fine-tuning models with differential privacy (DP) \cite{yue-etal-2023-synthetic, ramesh-etal-2024-evaluating, mattern-etal-2022-differentially} over sensitive corpora, others have developed methods that implement differential privacy during model inference \cite{wu2023privacypreservingincontextlearninglarge, tang2024privacypreserving, nahid2024safesynthdpleveraginglargelanguage}. 

\begin{figure*}[!ht]
    \centering
    \includegraphics[width=\linewidth]{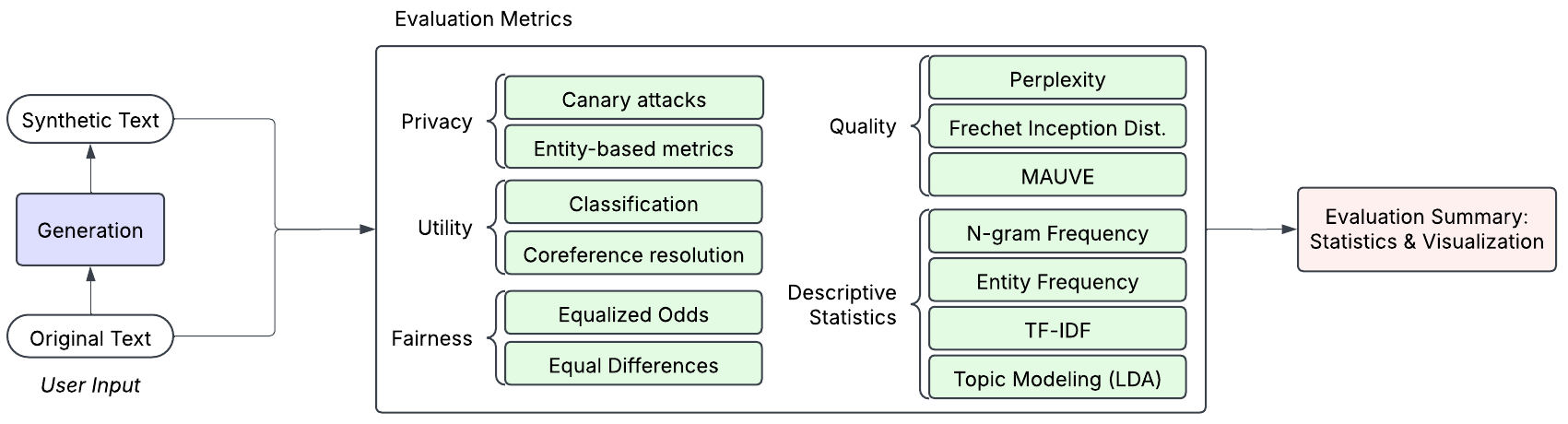}
    \caption{Architecture overview of \syntheval.}
    \label{fig:overview}
\end{figure*}

Despite substantial recent interest in privacy-preserving text generation \cite{yue-etal-2023-synthetic, mattern-etal-2022-differentially,wu2023privacypreservingincontextlearninglarge, tang2024privacypreserving, nahid2024safesynthdpleveraginglargelanguage},
evaluation criteria vary across studies, leading to inconsistencies in conclusions and results that overlook key aspects of data, such as fluency and privacy \cite{ramesh-etal-2024-evaluating}. Existing frameworks for assessing synthetic text quality either lack a wide range of criteria and metrics or do not provide easy-to-use tools to evaluate one's own data \cite{belgodere2024auditing,chim2024evaluating}. Given the importance of synthetic data generation and its potential use in high-stakes domains, there is a critical need for a thorough and consistent evaluation approach that supports comparability across studies. To this end, we present \syntheval{}: an open-source toolkit\footnote{GitHub Repository: \url{https://github.com/kr-ramesh/synthtexteval}} for evaluating synthetic text and comparing generation methods. An overview of the architecture is provided in Figure \ref{fig:overview}. While generalizable to many synthetic text use cases, our toolkit particularly aims to enable the promising potential synthetic text has for facilitating privacy-preserving data sharing in high-stakes domains.

\syntheval{} has three core features that support this endeavor. First, it provides functionality to generate synthetic text with differentially private  guarantees. Second, it implements a principled suite of generalizable metrics for evaluating the utility, privacy, fairness and quality of synthetic text, along with a descriptive outline of the data distribution.
Third, a user-friendly GUI enables manually comparing synthetic text with real text, supporting the solicitation of qualitative feedback from non-technical domain experts. 
We further provide built-in support for generating and evaluating synthetic data in two domains with sensitive data: law \cite{Pilán_Lison_Øvrelid_Papadopoulou_Sánchez_Batet_2022} and healthcare \cite{johnson2016mimic}, but our toolkit is crucially data-agnostic, allowing users to run our generalizable metrics over their own datasets.
Although our work is motivated by a clear need for standardized evaluation practices in high-stakes domains, it is broadly applicable to synthetic data for any task. By providing standardized data-agnostic metrics for evaluating synthetic data quality, we aim to assist researchers and practitioners in developing more robust and trustworthy AI systems and tools.

\section{Background} % and Related Work

Several factors influence the viability of synthetic data, including the faithfulness of the data to the source distribution, the diversity of the generated text, and its usability for developing downstream applications \cite{long-etal-2024-llms}.
% In Table \ref{tab:comparison-frameworks}, we summarize synthetic text evaluation metrics that have been used in recent studies.
Prior work has often used evaluation criteria targeted to specific use cases, which make synthetic data generation methods difficult to compare.
For instance, in generating synthetic text for privacy preservation, studies report metrics targeting training data leakage \citep{yue-etal-2023-synthetic}, but do not assess if the synthetic data introduces unfairness. In contrast, a system that aims to use synthetic data to address data imbalance may report results from groupwise performance metrics, but not privacy leakage \cite{pereira2024assessment}. Although not every metric is relevant for every use case, many of them do intersect: synthetic data is not a viable solution for privacy preservation if it introduces unfairness.

Even when the specified criterion is shared (e.g., privacy), studies often use different evaluation metrics %(Table \ref{tab:comparison-frameworks})
, which can be contradictory and preclude comparability of methods \cite{friedler2021possibility}. Motivated by the need for consistent evaluation, our framework implements evaluation metrics that have been commonly used in prior work, but lack standardization (e.g., classification, canary attacks), as well as introduces metrics that have been shown to identify weakness in synthetic data (e.g., coreference resolution, fairness criteria) but are rarely reported \citep{ramesh-etal-2024-evaluating}. 
 
 % While some criteria (such as those related to membership inference in privacy) already are widely used, we focus on what measures would be most useful to practitioners working with synthetic data.

\newcommand{\greencheck}{\textcolor{forestgreen}{\checkmark}}
\newcommand{\brickx}{\textcolor{brickred}{\ding{55}}}

 \begin{table*}
 \centering
 \scriptsize
 \begin{tabular}{ccccccccc}
 \toprule
  & \multicolumn{5}{c}{Evaluation Metrics} & \multicolumn{2}{c}{Ease-of-Use} & \multicolumn{1}{c}{Modality} \\ \cmidrule(lr){2-6} \cmidrule(lr){7-8} \cmidrule(lr){9-9} 
  & Utility & Privacy & Fairness & Quality & Descriptive & Code Package & GUI & Natural Language \\ \midrule
\syntheval & \greencheck & \greencheck & \greencheck & \greencheck & \greencheck & \greencheck & \greencheck & \greencheck \\
\citet{belgodere2024auditing} & \greencheck & \greencheck & \greencheck & \greencheck & \greencheck & \brickx & \brickx & \greencheck \\
\citet{chim2024evaluating} & \greencheck & \greencheck & \brickx & \greencheck & \brickx & \brickx & \brickx & \greencheck \\
\citet{gehrmann2021gem} & \greencheck & \brickx & \brickx & \greencheck & \brickx & \greencheck & \greencheck & \greencheck \\
\citet{SDV} & \greencheck & \greencheck & \greencheck & \greencheck & \greencheck & \greencheck & \greencheck & \brickx \\
\citet{synthcity} & \greencheck & \greencheck & \greencheck & \greencheck & \greencheck & \greencheck & \greencheck & \brickx \\ \bottomrule
 \end{tabular}
 \caption{Comparison of evaluation frameworks for private synthetic text data generation.}
 \label{tab:comparison-sdg-eval-frameworks}
 \end{table*}

\subsection{Synthetic Data Evaluation Frameworks}

Our toolkit differs from existing frameworks for evaluation of synthetic text in its wide range of criteria and metrics and provision of easy-to-use tools to evaluate one's own synthetic text data. \citet{belgodere2024auditing} provides a framework for auditing synthetic data across modalities, including tabular, time series, and text, and across fidelity, privacy, utility, and fairness metrics. However, due to their broad focus across modalities, their metrics are less tailored to text data, basing the evaluation on  embeddings only. \citet{chim2024evaluating} provides a text-specific auditing framework, focusing on user-generated text in messaging and social media contexts. They include metrics for downstream utility, privacy, and text fidelity, but do not include descriptive statistics or fairness metrics. Neither framework provides a user-friendly code package nor a GUI to support qualitative evaluations.

In addition to synthetic text evaluations frameworks, other benchmarks, such as \citet{gehrmann2021gem}, focus on model capabilities (e.g., developed on synthetic data) in the context of natural language generation tasks, but are less focused on metrics specific to the generated text. Complementary to our work, \citet{patel2024datadreamer} developed the DataDreamer tool to standardize workflows for generating synthetic text data with LLMs, providing functionality for both finetuning generative models and training downstream models on generated synthetic data. However, they do not provide any functionality to assess the generated synthetic data. \citet{SDV} and \citet{synthcity} both provide comprehensive evaluation frameworks and code packages for synthetic tabular data, but do not support evaluation of synthetic text.

\section{Design and Implementation}
\label{sec:design}
\syntheval{} contains a text generation module and a set of evaluation modules, for each category of evaluation metrics, and a GUI as shown in Figure \ref{fig:overview}. Modules can be run individually to focus on specific metrics or together for a full audit.  

The text generation model includes optional privacy-preserving guarantees for users who require them. It offers functionality for a descriptive analysis of any corpora, along with filtering functionality to remove low-quality synthetic text samples. In the evaluation module, the package includes tools to train and test downstream models on synthetic data, as well as automated evaluation of text quality, privacy and fairness metrics. The complete set of evaluation metrics are further explained in the following subsections. % Section \ref{sec:design}.

\subsection{GUI}

\begin{figure}
    \centering
    \includegraphics[width=0.9\linewidth]{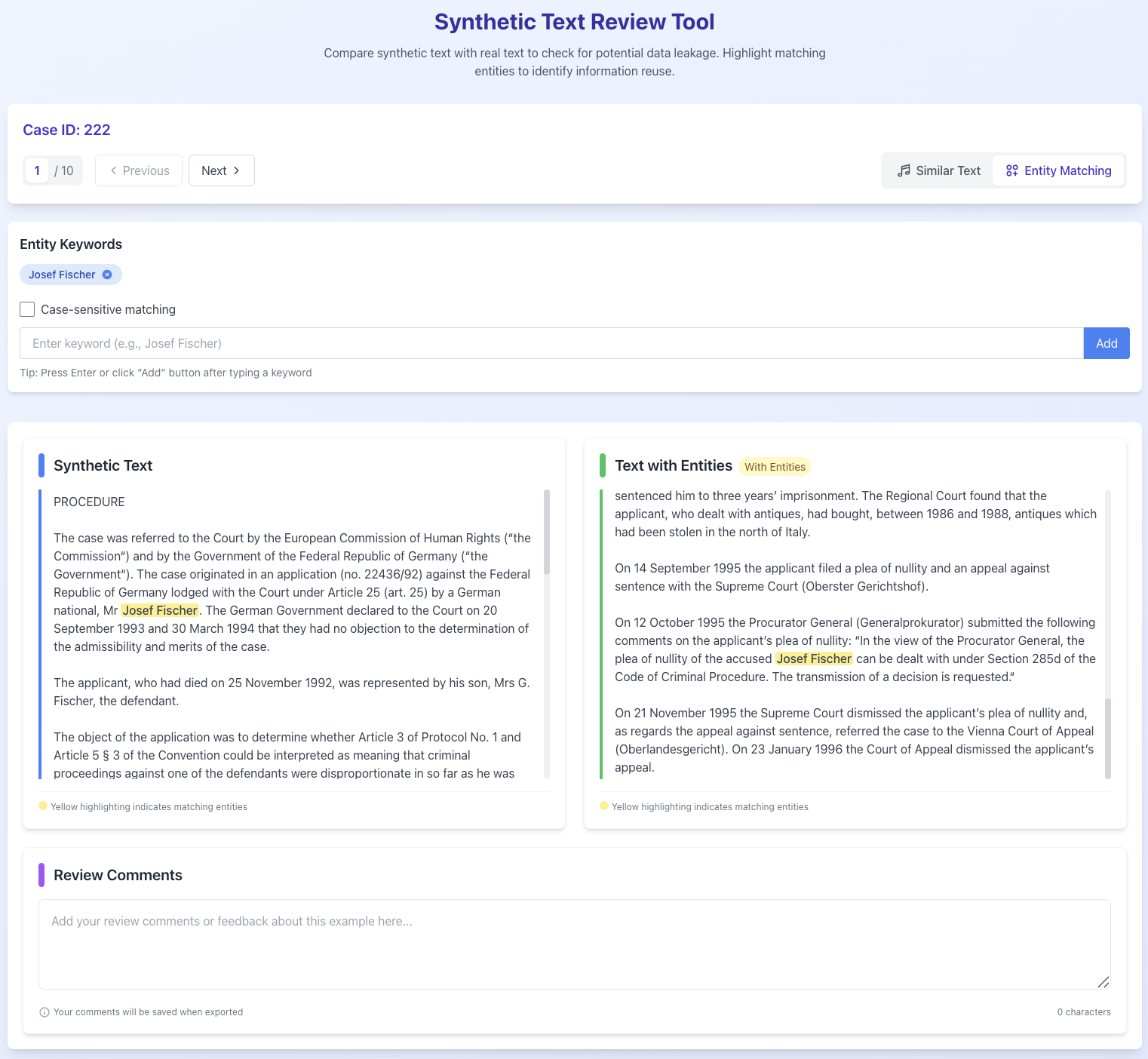}
    \caption{Our visual interface supporting exploration, comparison, and annotation of synthetic and real text.}
    \label{fig:gui}
\end{figure}

Although we implement comprehensive and generalizable evaluation metrics, fully automated metrics can miss more nuanced aspects of synthetic text quality, especially ones that may vary across domains. For example, leakage of a name of a judge in a legal text may be acceptable, whereas leakage of a patient name in clinical text is not.
To account for this, our framework contains a GUI that provides two key functionalities: i) for a given synthetic text, it displays the most real texts (determined through embedding similarity)  ii) for a given synthetic text that contains a named entity, it displays real texts containing the entity. The GUI further supports writing free-form comments on notes, which can be saved and shared. The GUI is shown in Figure \ref{fig:gui}.

This functionality in a user-friendly display allows non-technical domain experts to provide qualitative feedback on synthesized text, such as if synthesized text is factually accurate to real text and if it violates privacy by leaking information about entities.
% For instance, it can support a collaborative workflow where developers can share synthetic datasets and domain experts can review this data for potential privacy violations and assess the overall quality of the data to ensure that it aligns with application-specific requirements.

% \todo{Addign the link to the dataset}

\subsection{Installation and Setup}
\syntheval{} is an installable Python package\footnote{\url{https://pypi.org/project/SynthTextEval/}}, and the repository contains a number of user resources including detailed setup instructions, information about various components, as well as test functions for each of the package's modules. In addition, we provide a demo Jupyter notebook and runnable experiment scripts for our case studies. A web-based version of our GUI is available for interactive exploration\footnote{Online interface: \url{https://syntheticreview.cdhai.com/}}, and installation instructions for the desktop application are provided in the GitHub repository.

\subsection{Downstream Utility}

We assess the utility of synthetic data as a substitute for training models on real data with built-in support for two tasks: classification and coreference resolution. We select these two tasks both for their popularity in conducting evaluations (Table \ref{tab:comparison-sdg-eval-frameworks}) driven by potential real-word use cases \citep{gandhi-etal-2023-annotating}, and because the characteristics of the synthetic data that would increase its utility differs between the two tasks. For example, if the synthetic data successfully trains a high-performing classification model, this would suggest it contains a comprehensive set of features with similar frequency and contextual usage to the real data. However, lack of coherence or logical inconsistencies across synthesized documents may not be reflected in classification performance. In contrast, coreference resolution benefits from entity consistency, coherent structure, and the preservation of long-range dependencies.
% Thus, evaluating synthetic data based on its suitability for training models for both tasks provides a more comprehensive utility evaluation.

In practice, a researcher or practitioner might use synthetic text as training data for these tasks by hiring annotators to label it, but collecting these ``gold'' annotations is too time-consuming and expensive to evaluate synthetic data quality across generation methods. Instead, for both tasks, we offer the functionality to generate ``silver'' annotations for the synthetic data by using an external pretrained model. New classification and coreference models can then be trained on this silver-annotated data, with the models' performance evaluated on a (real) gold standard test set. For classification, we focus on both F1 score and accuracy and for coreference resolution we focus on F1 score. %The modularity of our framework enables users the seamless integration of other downstream tasks such as summarization and question-answering, independent of other evaluation modules.

%To evaluate the utility of synthetic text, we first compare the performance of downstream text classification models trained on real versus synthetic (or synthetically augmented) datasets. Augmentation of training datasets is one of the most common uses of synthetic text, as increasingly larger deep learning models require larger training datasets, and the focus on new tasks leads to small ``real'' datasets \cite{feng2021survey}. Replacement of the original dataset with synthetic data is relevant when in contexts where privacy is essential, and the original dataset cannot be shared (or used in downstream model training due to risks of leakage).

\subsubsection{Fairness}

\syntheval{} provides functionality to evaluate fairness in models trained on synthetic data. In domains such as healthcare and social services, the development of equitable systems is essential \cite{Meng2022InterpretabilityAF, 10.1145/3593013.3594094}. Prior research has identified differences in text data associated with difference groups, such as disparate rates of stigmatizing language in clinical notes \cite{harrigian2023characterization}, and examples where synthetic data increases downstream performance more for certain groups than others \cite{bhanot2021problem}. Thus, for the tasks where we conduct utility evaluations, we additionally assess whether the distribution of utility is even across designated subgroups.
We implement two commonly used fairness metrics: equalized odds (EO) and equality difference (ED). % Equalized odds measures whether the quality of the predictions are independent of the subgroup label, specifically, differences in the false positive rate (FPR) and true positive rate (TPR).
% Equalized difference sums the differences between subgroup performance and overall performance for a specific metric, e.g., false negative rate (FNED).
We provide full formulas in Appendix \ref{app:fairness_metrics}.

\subsection{Privacy}

Our toolkit supports privacy evaluations for language models using both canary-based evaluations \cite{10.5555/3361338.3361358} and entity-centric metrics \cite{ramesh-etal-2024-evaluating}. Out of proposed methods to quantify privacy leakage and memorization in LLMs \cite{lm-memorization, schwarzschild2024rethinking, kim2023propile, huang-etal-2024-privacy, li-etal-2024-privlm}, we focus specifically on measures that estimate the frequencies or likelihood of generation of spans of sensitive information and personal identifiable information (PII) in the synthetic text. These methods evaluate privacy risks in generated text, rather than vulnerability of models to attack (e.g., membership inference attacks), making them more suitable for evaluating synthetic text and also generalizable to diverse paradigms of privacy-preserving text generation, like sanitization approaches \citep{chen-etal-2023-customized}. 

Canary-based evaluations measure memorization through the generation likelihood of injected canary phrases. While they offer a controlled way to assess data leakage \cite{yue-etal-2023-synthetic}, they are not always reflective of actual PII leakage \citep{ramesh-etal-2024-evaluating}.  Thus, we additionally implement entity-centric evaluation criteria, which estimate the frequency of sensitive entities and their contexts in synthetic text. Appendix \ref{app:privacy_metrics} provides additional details.

%We propose two metrics to measure the privacy leakage of the model. First, the Private Information Package Percentage (PIPP) quantifies the ratio of synthetic texts containing private information to the total number of generated passages. Second, the Entity Leakage Percentage (ELP) measures the proportion of private entities that are leaked. To identify private entities within the text, we employ an entity-matching algorithm to extract and track these entities throughout the generated content.

\subsection{Text Quality}

%However, automated evaluation metrics have known limitations, as demonstrated in \todo{cite Blind Spots}, where stress-testing these metrics revealed weaknesses in their robustness across different test cases that correspond to specific types of text. Nevertheless, we chose to integrate an automated evaluation suite in our framework as these metrics often correlate with human judgment and can be useful indicators of text quality, especially when human evaluation is impractical due to resource constraints. 

In order to evaluate the overall synthetic text quality, independently from specific downstream utility tasks, our toolkit implements a range of measures, including general perplexity-based evaluations as well as the reference-based metrics Fr\'echet Inception Distance (FID) \cite{10.5555/3295222.3295408} and MAUVE \cite{pillutla2021mauve}.
These metrics have been shown to often align with human judgment and can serve as valuable indicators of text quality, especially when human evaluation is impractical due to resource constraints. While these metrics have known `blind spots' \citep{he-etal-2023-blind}, using a combination of automated qualitative metrics can offer useful insight into the distributional similarity between synthetic and real data.

FID calculates the feature-wise mean and covariance matrices of embeddings for both synthetic and real text, and the Fr\'echet distance between these matrices serves as a measure of their similarity. MAUVE is a KL-divergence-based metric that compares the distributions of the synthetic and real text in the embedding space of a selected LLM. %, accounting for scenarios where models either assign high probabilities to unrealistic sequences or struggle to replicate fluent, human-like text. 
MAUVE's strength lies in its usefulness for relative comparisons regardless of embedding model choice, as it is robust to scaling, quantization, and embedding choices, while also correlating strongly with human judgment compared to other automated metrics. Perplexity assesses text quality by measuring the likelihood that a provided base LLM would generate a given text segment. A higher perplexity score indicates greater semantic coherence \cite{colla2022semantic}. Further details on how each metric is calculated are in Appendix \ref{sec:qual-metrics}.

\subsection{Descriptive Statistics}

Users can conduct an exploratory text analysis of the synthetic data and make comparisons with real data by using the descriptive statistics module. Some of the functions included are an n-gram frequency analysis to detect common collocations and phrase patterns, TF-IDF computations to assess the relative importance of words across documents and an analysis of the most and least frequent entities in the text distributions. Additionally, to extract underlying thematic structures, the package employs Latent Dirichlet Allocation (LDA) for topic modeling, showing the user dominant topics and their associated keywords. 

%We additionally provide a visualization dashboard that summarizes synthetic text data evaluations across metrics. As shown in Figure \ref{}, the top-left quadrant displays downstream utility metrics, the top-right quadrant shows fairness metrics, the bottom-left quadrant displays privacy metrics, and the bottom-right quadrant shows descriptive metrics.

\subsection{Generating synthetic text}

Although our main focus is on facilitating evaluation, we additionally implement a popular synthetic data generation method, which involves fine-tuning an LLM with control codes, with or without differential privacy \cite{keskar2019ctrl,yue-etal-2023-synthetic,ramesh-etal-2024-evaluating}. This functionality allows users to conduct an initial assessment of synthetic data quality in their own domains, as well as provides a baseline to compare against more customized generation methods.

Control codes are a special type of prefix in the input to the language model that contain labels corresponding to the features associated with the desired synthetic text  \citep{keskar2019ctrl}. These control codes are prepended to the prompt during training, allowing the model to learn the relationship between the labels and the corresponding output distributions of the synthesized text. % (see Equation \ref{eq:control}). 
We provide example control codes in the appendix \ref{app:control-code-example}. During inference, users can supply their own control codes to the model, guiding it to generate text that matches the characteristics they desire for the intended use of the synthetic data.

% \begin{equation}\label{eq:control}
% P(x|c) = \prod_{i=1}^n P(x_i|x_1\dots x_{i-1}, c)
% \end{equation}

We incorporate support for differentially private (DP) fine-tuning of models to generate synthetic text with privacy guarantees. DP techniques safeguard the participants in a dataset by making it difficult to infer whether any single individual's data was used. This is done by injecting carefully calibrated random noise to the results of any analysis, so that the outcome is nearly the same whether or not a given data point is included. For synthetic text generation, we achieve these privacy guarantees by fine-tuning our models with DP-SGD (Differentially Private Stochastic Gradient Descent) \cite{Abadi_2016}. In DP-SGD, model gradients are clipped, and random noise is added during training, which limits the impact of any single data point and helps prevent the model from leaking sensitive information in its outputs. Further details about how DP works and its implementation in our study are provided in Appendix \ref{app:dp_background}.

While DP-based approaches offer well-defined upper bounds on the likelihood privacy leakage, this comes at the expense of model utility. Nevertheless, DP-generated synthetic text has been shown to be useful in reducing privacy risks while maintaining utility in certain downstream settings \cite{ramesh-etal-2024-evaluating, yue-etal-2023-synthetic}.

%DP provides an upper bound on the probability of privacy leakage, and reduces the risk of private information being regurgitated in synthetic outputs, although it comes at the expense of some utility. % Specifically, differential privacy determines guarantees on whether an individual's presence in a dataset ($D$) can be discovered through a query on that dataset ($F: D \rightarrow Y$) and a dataset that does not include that individual, $D'$. Given a privacy budget, $\epsilon$, a higher budget indicates an increased probability that an individual can be discovered (less privacy).
%
% \begin{equation}
% \text{Pr}[F(D) \in y] \le exp(\epsilon) \text{ Pr}[F(D') \in y]
% \end{equation}

\section{Validation and Example Usage}

\subsection{Experimental setup and objectives}

Our first case study uses the Text Anonymization Benchmark (TAB), a dataset of 1,268 European court cases \cite{tab_dataset} extensively annotated to support evaluating text anonymization methods. We generate two synthetic datasets: one without DP ($\epsilon=\infty$) and one with DP ($\epsilon=8$). We use the country and year of the court case as control codes for generation. Thus, our artificially created classification task is to predict the country for each court case. A BERT$_{base}$ model is finetuned on each of the datasets. Our privacy evaluation focuses on the reoccurrence of personal information entities from the original dataset in the synthetic datasets.

Our second case study uses clinical notes from MIMIC-III  \cite{johnson2016mimic} and coreference-annotated clinical notes from the i2b2/VA Shared-Task \cite{uzuner2012evaluating}. 
% MIMIC-III contains  $>2$M deidentified clinical notes, and the i2b2/VA dataset includes 251 documents annotated for coreference resolution.
For the MIMIC-III data, we use the 10 most frequent 
International Classification of Diseases (ICD-9) codes as the control codes to generate synthetic data (similar to \citet{al2021differentially}), also with $\epsilon=\infty$ and $\epsilon=8$.

To measure downstream utility, we evaluate performance on the multilabel and multiclass ICD-9  code prediction task for MIMIC-III, and the coreference resolution task for the i2b2/VA dataset. We compare performance for a BERT$_{base}$ model finetuned on only the original dataset to a model finetuned on the synthetic datasets. The fairness metrics for the ICD-9 code prediction task are calculated for both patient race and patient gender on the real clinical notes, while the entity-based privacy evaluations focus on the regurgitation of PII from the real data and the contexts in which they appear.

% and focuses on ICD-9 diagnosis code identification  \cite{huang2019empirical, edin2023automated, li2020icd} as the downstream task.

\subsection{Results}

We report classification utility for both datasets in Table \ref{tab:classification} and remaining results in Appendix \ref{app:results_tables}. In general,  results follow the trends we expect, where real text and non-DP synthetic text have the best utility and quality, but DP synthetic text has the best privacy.
These results validate our toolkit's ability to reflect these properties.

\subsubsection{TAB}

Descriptive statistics show the synthetic data for TAB is on average about half as long and has half as many unique words (Table \ref{tab:res-descriptive}), although this is influenced by the user's choice of hyperparameters in generating the data. The synthetic data without DP ($D_{\epsilon=\infty}$) has higher Jaccard Similarity and Cosine Similarity to the real dataset compared to the synthetic data with DP ($D_{\epsilon=8}$). Regarding quality, both synthetic datasets are on par with the real test set for FID, but are lower quality according to the MAUVE and perplexity metrics (Table \ref{tab:tab_quality}).

Utility results (Table \ref{tab:classification}) show the downstream model trained on $D_{\epsilon=\infty}$ has higher F1 and higher accuracy than the one trained on $D_{\epsilon=8}$. However, $D_{\epsilon=\infty}$ has a higher percent of PII leaked from the real data, indicating a higher privacy risk (Table \ref{tab:entity_phrase}). 
% Overall, our evaluation of our synthetic TAB data shows high downstream utility and moderate text quality, but relatively poor privacy preservation. 
There is relatively high privacy risks for both synthetic datasets, likely due to the small size of the original TAB dataset. The $D_{\epsilon=8}$ dataset trades off slightly better privacy metrics for lower quality and utility compared to $D_{\epsilon=\infty}$.

\begin{table}[h]
\centering
\resizebox{\columnwidth}{!}{
\begin{tabular}{cccc}
\hline
 & & \multicolumn{1}{c}{\textbf{F1 Score}} & \multicolumn{1}{c}{\textbf{Accuracy}} \\ \hline
 \hline
\multirow[c]{2}{*}{\textbf{TAB}} & $D_{\epsilon = \infty}$      & 0.96 ± 0.035 & 0.99 ± 0.012 \\
& $D_{\epsilon = 8}$     & 0.54 ± 0.0 \da{-0.42} & 0.95  ± 0.0 \da{-0.04} \\ 
\hline
\multirow[c]{3}{*}{\textbf{Healthcare}} & $D_{real}$      & 0.67 ± 0.012 & 0.32 ± 0.016 \\
& $D_{\epsilon = \infty}$      & 0.61 ± 0.003 \da{-0.06} & 0.26 ± 0.004 \da{-0.06} \\
& $D_{\epsilon = 8}$     & 0.40 ± 0.004 \da{-0.27} & 0.18 ± 0.005 \da{-0.14} \\ 
\hline
\end{tabular}}
\caption{\textbf{TAB}: Difference in performance between models trained on data generated with DP and models trained on data generated without DP over TAB classification. \textbf{Healthcare}: difference in performance between real and synthetic data as training data for the ICD-9 classification task.}
\label{tab:classification}
\end{table}

\subsubsection{Healthcare}

For both coreference resolution and mention detection, training on either synthetic dataset results in significantly lower utility compared to the real data, with $D_{\epsilon=\infty}$ and $D_{\epsilon=8}$ providing nearly identical performance (Table~\ref{tab:coref}). In contrast, for the ICD-9 classification task (Table~\ref{tab:classification}), there is a more pronounced drop in utility when using the differentially private synthetic data ($D_{\epsilon=8}$), while $D_{\epsilon=\infty}$ achieves higher performance than $D_{\epsilon=8}$ but still does not reach the performance of the model trained on the original dataset.

Fairness metrics (Table \ref{tab:mimic_fairness}) suggest the model trained on the real data is most fair for race as the sensitive attribute, with the lowest FNED and EO metrics. $D_{\epsilon=\infty}$ is second highest, and $D_{\epsilon=8}$ has the worst fairness. For gender as the sensitive attribute, the models are all approximately equally fair across training datasets. On the privacy criteria, $D_{\epsilon=\infty}$ includes a higher percent of reproduced entities compared to $D_{\epsilon=8}$. Overall, our clinical notes evaluation shows that generating data without DP provides higher downstream utility and fairer downstream models at the cost of slightly higher privacy risks, compared to synthetic clinical notes generated with DP.

\paragraph{Conclusion}
\syntheval{} provides a toolkit to comprehensively evaluate synthetic text data across a wide range of metrics. This toolkit allows users the opportunity to assess the relative merits of synthetic data generation approaches, through the standardization of evaluation metrics, allowing for greater confidence in synthetic data quality, especially in high-stakes domains where quality synthetic data is in high demand.

\section{Ethics Statement}

\syntheval{} is designed to lower barriers to development of AI systems in sensitive domains through comprehensive synthetic text data evaluation. Rather than rely on ad-hoc evaluation and comparison of synthetic text data, we hope our toolkit will help standardize comparisons and encourage responsible synthetic text data generation. In particular, our fairness and privacy metrics emphasize that high performance on utility and quality metrics are not on their own sufficient criteria for determination of worthiness for synthetic data publication and distribution.

However, while we have aimed to include wide range of metrics across many criteria, we caution that other context-dependent assessment may be necessary.An unintended consequence of this package could be a false sense of security from satisfactory performance on the included metrics, leading to less consideration of other context-specific evaluations. Additionally, there may be contexts where it is unethical or harmful to generate synthetic text data, regardless of performance on evaluations.
%For example, stigma in generated clinical notes may be important to measure prior to release, but is not currently included. 
We are committed to continuing to improve this package. Future efforts will focus on widening the evaluation criteria and metrics included, providing further visual or interactive tools for comparison, and enhancing reliability of the package.

\section*{Acknowledgments}
We thank reviewers for their helpful feedback. This work was supported in part by the AI2050 Fellowship program by Schmidt Sciences. We also thank Tianli Xu for his valuable contribution in developing the annotation interface for this paper.

\bibliography{anthology,custom}

\clearpage

\appendix

\section{Background: Differential Privacy}
\label{app:dp_background}

Differential privacy (DP) offers a formal privacy guarantee that ensures that any individual's data cannot be inferred from a query applied to a dataset \citep{Dwork2006,dwork2014algorithmic}. In other words, the result of such a query is nearly indistinguishable from the result of the same query applied to a dataset that either includes a modified version of the individual's data or excludes the record entirely, thereby preserving the individual's privacy. In this case, the notion of adjacency is defined as a difference of a single record in the original dataset D and the modified dataset D'. Formally, DP is defined as follows:
%There has been prior discussion on the specificities and variants of differential privacy (DP) within NLP \citep{Lyu2020}.

\textbf{Definition}: Given a dataset \(D\) and an adjacent dataset \(D'\), which is produced by removing or modifying a single record from \(D\), a randomized algorithm \(F : D \to Y\) is \((\epsilon, \delta)\)-private if for any two neighboring datasets \(D, D'\), with the constraints \(\epsilon > 0\) and \(\delta \in [0, 1]\), the following holds true for all sets \(y\) \(\subseteq\) \(Y\):
\[
\Pr[F(D) \in y] \leq e^{\epsilon} \Pr[F(D') \in y] + \delta
\]

The value of \(\epsilon\) denotes the privacy budget, while \(\delta\) specifies the likelihood that the privacy guarantee may fail. When \(\delta\) is set to 0, this implies a purely differentially private setting with no probability of the guarantee being broken. The value of \(\epsilon\) constrains how similar the outputs of both distributions are; a higher \(\epsilon\) value indicates a greater privacy budget, meaning the algorithm is less private. DP guarantees that even if an adversary has access to any side-knowledge, the privacy leakage of \((\epsilon, \delta)\)-DP algorithms will not increase. Additionally, another property of DP is that it ensures that any post-processing on the outputs of \((\epsilon, \delta)\)-differentially private algorithms will remain \((\epsilon, \delta)\)-differentially private. 

We use DP-SGD \citep{Abadi_2016}, a modification to the stochastic gradient descent (SGD) algorithm, which is typically used to train neural networks. DP-SGD clips the gradients to limit the contribution of individual samples from the training data and subsequently adds noise from a predefined type of distribution (such as a Gaussian distribution) to the sum of the clipped gradients across all samples. DP-SGD thus provides a differentially private guarantee to obfuscate the gradient update, thereby ensuring that the contribution of any given sample in the training data is indistinguishable due to the aforementioned post-processing property. This process ensures \((\epsilon, \delta)\)-differential privacy for each model update. Given a privacy budget, number of epochs, and other training parameters, we can estimate the privacy parameters using estimation algorithms \cite{gopi2021numerical}.

\section{Privacy metrics}
\label{app:privacy_metrics}

%Our toolkit offers functionality for canary-based evaluations \cite{10.5555/3361338.3361358}, along with functions to quantify the Entity Leakage Percentage (ELP) and the proportion of an entity's context leaked, where the user can define the context window length \cite{ramesh-etal-2024-evaluating}. Out of proposed methods to quantify privacy leakage and memorization in LLMs \cite{lm-memorization, schwarzschild2024rethinking, kim2023propile, huang-etal-2024-privacy, li-etal-2024-privlm}, %although no unified benchmark or toolkit exists at present that consolidates all of these approaches. 
%we focus specifically on measures that estimate the frequencies or likelihood of generation of spans of sensitive information and personal identifiable information (PII) in the synthetic text.
%These methods focus on evaluating risks of privacy leakage in generated text, rather than vulnerability of models to attack (e.g., membership inference attacks), making them more suitable for evaluating synthetic text quality and also generalizable to diverse paradigms of privacy-preserving text generation, like sanitization approaches \citep{chen-etal-2023-customized}.

Canary-based evaluations \cite{10.5555/3361338.3361358} involving crafting canaries that contain sensitive information and injecting them into the training data. The probability of generating the canary is estimated in contrast to the model trained on the corpus without the canary (a model that has ``memorized'' the canary would have a higher likelihood of generating it). In addition to this, we implement entity-centric evaluation criteria to quantify the Entity Leakage Percentage (ELP) and the proportion of an entity's context leaked, where the user can define the context window length \cite{ramesh-etal-2024-evaluating}.

The ELP is calculated as follows, where $E_{\text{leaked}}$ is the number of entities leaked in the synthetic generations, and $E_{\text{total}}$ represents the total number of entities in the training data.
\begin{equation*}
P_{\text{entities}} = \frac{E_{\text{leaked}}}{E_{\text{total}}} \times 100
\end{equation*}

Similarly, we quantify the memorized spans of context where the entities appear as follows: 
\begin{equation*}
%P_{\text{context}}(k) = \frac{C_{\text{leaked}}(k)}{C_{\text{total}}(k)} \times 100,       
P_{\text{context}}(k) = \frac{C_{\text{leaked}}(k)}{C_{\text{total}}(k)} \times 100
\end{equation*}

where $C_{\text{leaked}}(k)$ is the number of occurrences where the entity and its surrounding context (of length $k$) are leaked in the synthetic generation and $C_{\text{total}}(k)$ is the total number of occurrences where the entity and its context (of length $k$) appear in the original training data.

\section{Fairness metrics}
\label{app:fairness_metrics}
Where $D$ is a set consisting of all subgroups corresponding to a demographic attribute within the dataset, Equalized Odds is defined as:

\begin{multline*}
\text{EO}_{D} = \max \Bigl( 
\max_{d \in D}(\text{TPR}_d) - \min_{d \in D}(\text{TPR}_d), \\
\max_{d \in D}(\text{FPR}_d) - \min_{d \in D}(\text{FPR}_d) 
\Bigr)
\end{multline*}

Equalized difference sums the differences between subgroup performance and overall performance for a specific metric. For example, FPR is an important metric in domains such as healthcare, where false positive diagnoses can place increased burdens on patients in the form of unnecessary treatment or exams \cite{rajkomar2018ensuring}. Thus, the False Positive Equality Difference (FPED) metric is the sum of the differences between the overall FPR for the entire dataset and the FPR for each subgroup:

\begin{equation}
\text{FPED} = \sum_{d=1}^{D} \left| \text{FPR}_{\text{overall}} - \text{FPR}_{d} \right|
\end{equation}

ED metrics for true positive (TPED), true negative (TNED) and false negative (FNED) are computed similarly. Lower values of EO and ED scores indicate that the model's performance is more consistent across different subgroups.

\section{Automated Text Quality Evaluation Metrics}

\label{sec:qual-metrics}

We provide the full equations to calculate our two automated qualitative evaluation metrics, MAUVE and Fr\'echet Inception Distance.

MAUVE is computed by first sampling human text $x_i \sim P$ (reference) and generated text $x_i' \sim Q$ (synthetic). Embeddings for this text are then obtained from an external model M (we use GPT-2 \cite{}). These embeddings are quantized through clustering, yielding ``low-dimensional discrete distributions that approximate each
high-dimensional text distribution'' \cite{pillutla2021mauve}. Finally, a divergence curve is calculated by taking the KL-divergence between the reference distribution and synthetic distribution as a mixture weight $\lambda$ is varied, as shown in Equation \ref{eq:mauve}, and MAUVE is calculated as the area under this curve.
%
%\begin{multline}\label{eq:mauve}
%C(P,Q) = \Bigl\{(exp(-c\text{ KL}(Q|R_\lambda)), \\
%exp(-c\text{ KL}(P|R_\lambda)) : \\
%R_\lambda = \lambda P + (1-\lambda) Q, %\lambda \in (0,1)\Bigr\}
%\end{multline}

\begin{multline}\label{eq:mauve}
C(P, Q) = \Bigl\{ \bigl(e^{-c \text{KL}(Q \| R_\lambda)}, e^{-c \text{KL}(P \| R_\lambda)}\bigr) : \\
R_\lambda = \lambda P + (1-\lambda) Q, \, \lambda \in (0,1) \Bigr\}.
\end{multline}

The Fr\'echet Inception Distance is obtained by measuring the Wasserstein-2 distance between the Gaussian obtained from the reference data $(m, C)$ and the Gaussian of the synthetic data $(m_w, C_w)$:

\begin{multline*}
d^2((m,C), (m_w, C_w)) = \|m-m_w\|_2^2 + \\
Tr(C + C_w - 2(CC_w)^{1/2}
\end{multline*}

\section{Control Codes}\label{app:control-code-example}

We adapt the dominant approach from prior work \citep{yue-etal-2023-synthetic}: which involves first fine-tuning a pre-trained autoregressive language model on real, in-domain data and then generating synthetic data from the fine-tuned model.
We compare fine-tuning the model with and without DP, where we use DP-SGD for differentially private fine-tuning. After fine-tuning, we utilize top-k sampling \cite{fan-etal-2018-hierarchical} and nucleus sampling \cite{nucleus} to generate diverse synthetic notes.

We condition the text generation on \textit{control codes} \citep{keskar2019ctrl}. During training, we prepend one or more labels associated with the text to the model input. We similarly prepend control codes during inference, where we sample the provided codes from their distribution in the training data. Thus, during training and inference, the probability distribution of the subsequent text $x = \{x_{1}, x_{2} ... x_{n}\}$ is conditioned on the control code information $c$, which is described by the following equation:

\begin{equation}
P(x|c) = \prod_{i=1}^{n} P(x_i |x_{1} ... x_{i-1}, c)
\end{equation}

The format of the control code for the MIMIC-III data is as follows: \textit{Long\_Title: <diagnoses>, ICD9\_CODE: <codes>, Gender: <gender>, Ethnicity: <ethnicity>}, where the <diagnoses> variable represents the long title form of the ICD-9 codes, information that is already provided with the MIMIC-III dataset. An example control code is as follows: \textit{Diagnosis: Long\_Title: Unspecified essential hypertension, Atrial fibrillation ICD9\_CODE: 4019, 42731 Gender: Female Ethnicity: WHITE} 

\section{Sample Usage Results}
\label{app:results_tables}

\begin{table}[h]
\centering
\resizebox{\columnwidth}{!}{%
\begin{tabular}{@{}ccccccc@{}}
\toprule
\textbf{Domain} & \textbf{\begin{tabular}[c]{@{}c@{}}Avg Length\\ Of Dsynth\end{tabular}} & \textbf{\begin{tabular}[c]{@{}c@{}}Avg. Num\\ Unique Words\end{tabular}} & \textbf{\begin{tabular}[c]{@{}c@{}}Min Text\\ Length\end{tabular}} & \textbf{\begin{tabular}[c]{@{}c@{}}Max Text\\ Length\end{tabular}} & \textbf{\begin{tabular}[c]{@{}c@{}}Jaccard \\ Similarity\end{tabular}} & \textbf{\begin{tabular}[c]{@{}c@{}}Cosine \\ Similarity\end{tabular}} \\ \midrule
TAB (DP - inf)  & 617.725                                                                 & 265.882                                                                  & 452                                                                & 745                                                                & 0.185                                                                  & 0.522                                                                 \\
TAB (DP - 8)    & 613.75                                                                  & 245.95                                                                   & 8                                                                  & 787                                                                & 0.147                                                                  & 0.474                                                                 \\
TAB (Real)      & 1342.42                                                                 & 487.966                                                                  & 185                                                                & 5144                                                               & -                                                                      & -                                                                     \\ \bottomrule
\end{tabular}%
}
\caption{Descriptive statistics for TAB across real and synthetic datasets.}
\label{tab:res-descriptive}
\end{table}

\begin{table}[]
\centering
\resizebox{\columnwidth}{!}{
\begin{tabular}{ccccc}
\toprule
    & \textbf{FID} & \textbf{MAUVE} & \textbf{Avg. Perplexity}  \\ \hline\hline
    $D_{real (test)}$ & 0.788 & 0.941 & 15.38 \\
    $D_{\epsilon=\infty}$ & 0.797 & 0.485 & 11.474 \\
    $D_{\epsilon=8}$ & 0.790 & 0.831 & 10.209 \\
    \bottomrule
\end{tabular}}
\caption{Text quality metrics for TAB. The first column indicates the comparison dataset, with $D_{real(train)}$ as the reference dataset for each.}
\label{tab:tab_quality}
\end{table}

\begin{table}[h]
    \centering
    \begin{tabular}{lcccc}

\hline
& \multicolumn{1}{c}{Healthcare} &  & \multicolumn{1}{c}{TAB} \\
      \cline{1-2} \cline{3-4}
 & \textbf{\% Entities} & & \textbf{\% Entities} \\
 \hline
 \hline
 %$D_{real}$              &     1    &  216592  & & 1 & 2935955 \\
$D_{\epsilon = \infty}$    &  4.8 & &  50.3 \\
$D_{\epsilon = 8}$        &  3.8  & & 48.5 \\ \bottomrule
% $D_{\epsilon = 8}$        &  3.6  & &  - \\ 
    \end{tabular}
    \caption{Percent of unique PII from the training data that appear in the synthetic generations.}
    \label{tab:entity_phrase}
\end{table}

\begin{figure}
\includegraphics[width=\linewidth]{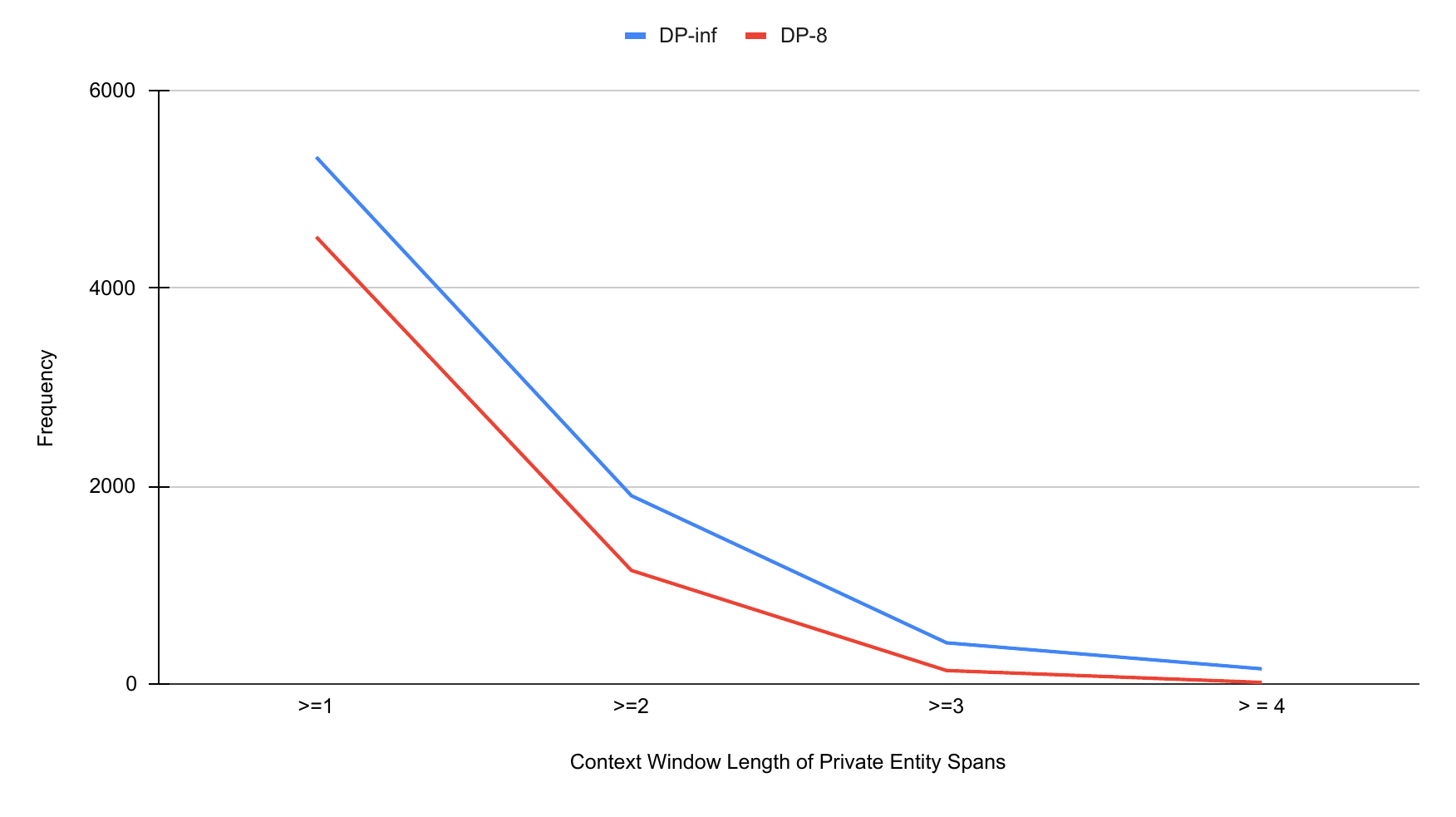}%
  \caption{Memorization of private entities in the TAB dataset as context window length increases.}
  \label{fig:entity_privacy_context_length}
\end{figure}

\begin{table}[]	
\centering
\resizebox{\columnwidth}{!}{
\begin{tabular}{lcc}
\hline				
& \textbf{Mention Detection} & \textbf{Coreference}\\\hline\hline		
$D_{real (gold)}$	&	0.800 ± 0.005	&	0.706 ± 0.006 \\
$D_{real (silver)}$	&	0.659 ± 0.121	&	0.552 ± 0.126 \\\hline
$D_{\epsilon = \infty}$	&	0.650 ± 0.008	&	0.483 ± 0.002 \\
$D_{\epsilon = 8}$	&	0.656 ± 0.006 &	0.491 ± 0.003	\\
\end{tabular}}	
\caption{F1 scores for coreference and mention detection for MIMIC-III over entities from human-annotated test splits of the i2b2/VA datasets. All synthetic datasets are annotated with silver labels.}
\label{tab:coref}	
\end{table}

\renewcommand{\arraystretch}{1.2}
\begin{table}[h]
\centering
\begin{tabular}{cccc}
\hline
                        & & \textbf{FNED}         & \textbf{Equalized Odds} \\ \hline
                        \hline
\multirow{3}{*}{\rotatebox[origin=c]{90}{\textbf{Race}}} &  $D_{real}$       & 0.35 ± 0.0 & 0.20 ± 0.0            \\
& $D_{\epsilon = \infty}$ & 0.39 ± 0.005 & 0.23 ± 0.001  \\
& $D_{\epsilon = 8}$               & 0.53 ± 0.014 & 0.30 ± 0.003  \\
\hline
\multirow{3 }{*}{\rotatebox[origin=c]{90}{\textbf{Gender}}}  & $D_{real}$       & 0.04 ± 0.0 & 0.04 ± 0.0            \\
& $D_{\epsilon = \infty}$ & 0.02 ± 0.007 & 0.02 ± 0.007           \\
& $D_{\epsilon = 8}$               & 0.04 ± 0.005 & 0.04 ± 0.005           \\
\hline
\end{tabular}
\caption{Fairness evaluation for the MIMIC-III $\text{ICD-9}_{n=10}$ task, for the gender and race categories. Higher values indicate poorer group fairness performance.}
\label{tab:mimic_fairness}
\end{table}

\begin{table}[th]
\small
\centering
\setlength{\tabcolsep}{8pt}
\begin{tabular}{lcc}
\toprule
 & \textbf{Rank} & \textbf{Perplexity} \\
 & $(\epsilon = \infty$\,/\,8) & $(\epsilon = \infty$\,/\,8) \\
\midrule
\multicolumn{3}{l}{\textbf{100 Insertions}} \\
Name    & 4628\,/\,3356 & 35.80\,/\,53.19 \\
Address & 5\,/\,3967    & 16.52\,/\,61.37 \\
Number  & 1\,/\,818     & 5.77\,/\,14.46 \\
Email   & 1\,/\,1410    & 10.50\,/\,70.26 \\
\addlinespace
\multicolumn{3}{l}{\textbf{10 Insertions}} \\
Name    & 5986\,/\,3378 & 49.72\,/\,54.10 \\
Address & 2276\,/\,4075 & 43.59\,/\,62.66 \\
Number  & 902\,/\,841   & 9.43\,/\,14.61 \\
Email   & 711\,/\,1452  & 37.81\,/\,72.08 \\
\addlinespace
\multicolumn{3}{l}{\textbf{1 Insertion}} \\
Name    & 6037\,/\,3383 & 52.06\,/\,54.20 \\
Address & 3348\,/\,4081 & 54.50\,/\,62.79 \\
Number  & 1084\,/\,838  & 9.82\,/\,14.63 \\
Email   & 1941\,/\,1457 & 43.90\,/\,72.28 \\
\addlinespace
\multicolumn{3}{l}{\textbf{0 Insertions}} \\
Name    & 6086\,/\,5265 & 44.81\,/\,57.54 \\
Address & 4565\,/\,3869 & 75.79\,/\,61.68 \\
Number  & 1217\,/\,1522 & 11.80\,/\,13.33 \\
Email   & 1003\,/\,3174 & 43.80\,/\,55.19 \\
\bottomrule
\end{tabular}
\caption{Rank and perplexity metrics for canary attacks over Healthcare (MIMIC) data. Each entry is shown as $\epsilon = \infty$\,/\,8. Differential privacy ($\epsilon=8$) increases both metrics, improving privacy for all canaries.}
\label{tab:canary_evaluations_app}
\end{table}

\end{document}